\documentclass{article}
\usepackage{ijcai25}

\usepackage{times}
\usepackage{soul}
\usepackage{url}
\usepackage[hidelinks]{hyperref}
\usepackage[utf8]{inputenc}
\usepackage[small]{caption}
\usepackage{graphicx}
\usepackage{amsmath}
\usepackage{amsthm}
\usepackage{booktabs}
\usepackage{algorithm}
\usepackage{algorithmic}
\usepackage[switch]{lineno}

\usepackage{bibentry}
\usepackage{xcolor}
\usepackage{colortbl}
\usepackage{tabularx} 
\usepackage{multirow}
\usepackage{enumitem}
\usepackage{rotating} 
\usepackage{makecell}
\usepackage{array}  

\title{Automated Decision-Making on Networks with LLMs through \\
Knowledge-Guided Evolution}

\author{
Xiaohan Zheng\and
Lanning Wei\and
Yong Li\and
Quanming Yao\thanks{Corresponding author}
\affiliations
Department of Electronic Engineering, Tsinghua University
\emails
zhengxh23@mails.tsinghua.edu.cn,
weilanning@163.com, 
liyong07@tsinghua.edu.cn, 
qyaoaa@tsinghua.edu.cn
}

\begin{document}

\maketitle

\begin{abstract}

Effective decision-making on networks often relies on learning from graph-structured data, 
where Graph Neural Networks (GNNs) play a central role, 
but they take efforts to configure and tune. 
In this demo, we propose LLMNet, 
showing how to design GNN automated through 
Large Language Models. 
Our system develops a set of agents that construct graph-related knowlege bases 
and then leverages Retrieval-Augmented Generation (RAG) 
to support automated configuration and refinement of GNN models 
through a knowledge-guided evolution process. 
These agents, equipped with specialized knowledge bases, 
extract insights into tasks and graph structures by interacting with the knowledge bases.
Empirical results show LLMNet excels in twelve datasets across three graph learning tasks, 
validating its effectiveness of GNN model designing.

\end{abstract}

\vspace{-0.3cm}
\section{Introduction and Related Work}
Effective decision-making in networks—such as in 
communication networks, social networks, and transportation networks—
often relies on graph-structured data representations. 
Among the techniques developed for learning from such data, 
Graph Neural Networks (GNNs) have become widely adopted across diverse domains, 
including tasks such as anomaly detection 
and recommendation systems in social networks~\cite{hamilton2017inductive}, 
as well as for predicting biomedical molecular properties~\cite{gilmer2017neural}.
The majority of existing GNNs are designed for diverse graphs under a specific task~\cite{wu2020comprehensive}, 
such as capturing graph-level representations~\cite{zhang2018end,ying2018hierarchical}.
and learning subgraph patterns in link-level tasks~\cite{he2020lightgcn,zhang2018link}.  
However, designing effective GNNs for different graph learning problems is challenging, 
as it requires substantial graph-related knowledge in order to understand the tasks and graphs~\cite{hoffman1995eliciting}.
Then, there is a natural question:
\textit{How to integrate graph learning knowledge to design effective GNNs?}
It is non-trivial to answer this question.
Firstly, existing methods have not provided explicit guidelines for utilizing knowledge in designing GNN model architectures.
Most GNNs are designed to effectively model graphs for a specific task~\cite{wu2020comprehensive,hamilton2017inductive,ying2018hierarchical}, 
based on implicit human expertise, which is difficult to explicitly describe and extract.

Therefore, we propose LLMNet, which automates GNN design using LLMs. 
Specifically, we have designed a Knowledge Agent to extract graph-related knowledge, 
building knowledge bases covering advanced graph learning research. 
Then, we have developed a set of agents that use RAG (Retrieval-Augmented Generation) 
to interact with knowledge bases, 
designing GNNs step by step in a knowledge-guided manner.
Leveraging LLMs' task analysis, LLMNet streamlines 
the designing and refinement of GNN model architectures.
Extensive experiments on twelve datasets across three tasks 
demonstrate LLMNet's superior performance and efficiency, 
proving the effectiveness of integrating knowledge for automated GNN design.
A concrete case demonstrating this process 
is presented in Section~\ref{sec-demo}.

\vspace{-0.2cm}
\section{Method}

We introduce LLMNet, which prepares and utilizes knowledge to design GNN model architectures for diverse graph learning tasks using LLM-based agents. 
Firstly, we gather graph-related resources and develop a knowledge agent for knowledge extraction and retrieval. 
Subsequently, the knowledge is then used by several LLM-based agents step by step to design effective GNN model architectures.

\begin{figure*}[t]
    \centering
    \includegraphics[width=0.95\textwidth]{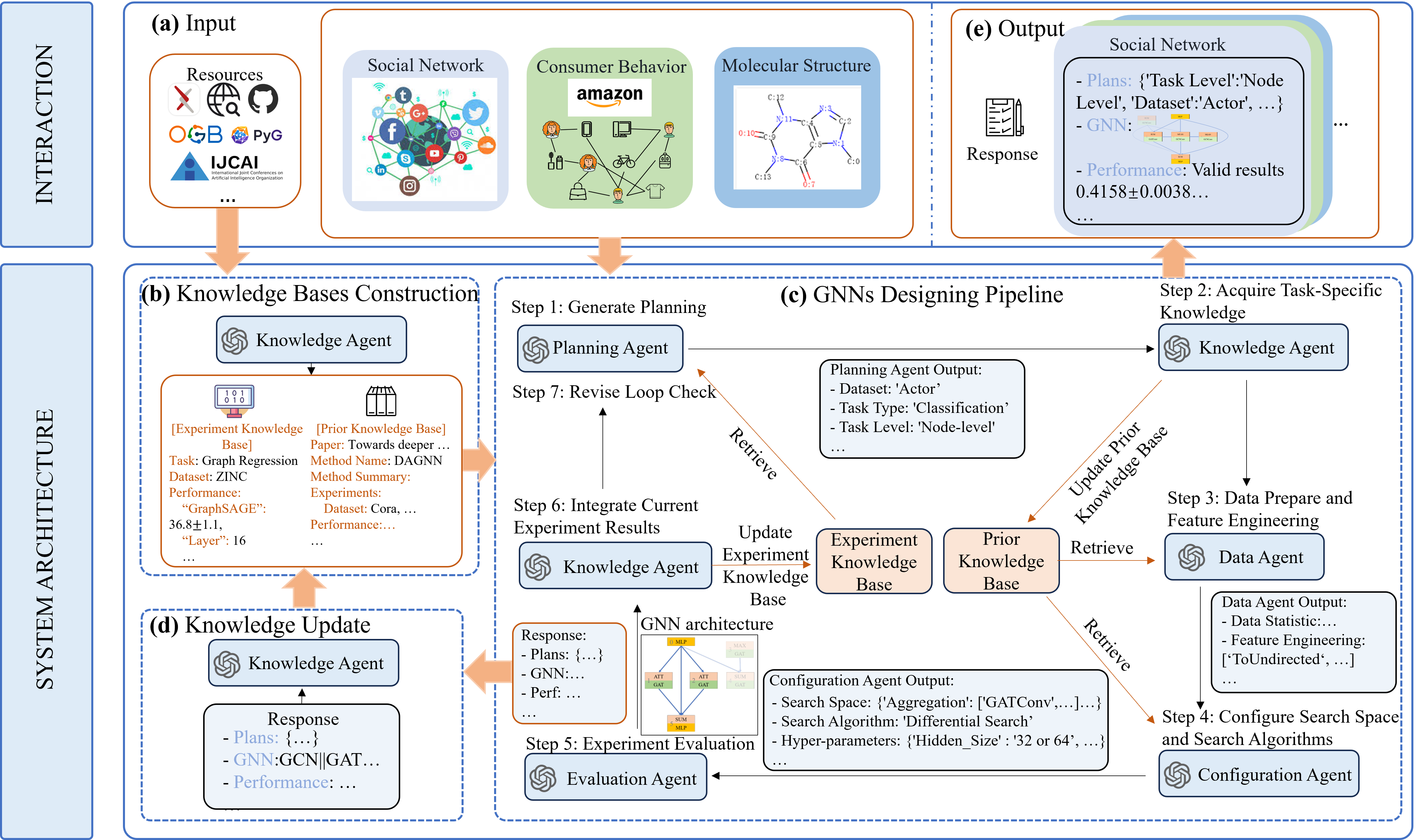} 
\caption{
System architecture of LLMNet. 
LLMNet is designed to automate the design of GNN model architectures through a knowledge-guided approach. 
(a) shows the system inputs: graph need to process, external resources for knowledge base construction.
(b) shows how the Knowledge Agent builds the knowledge bases. 
(c) depicts the automated GNN design and finetuning process, where the constructed knowledge is utilized by a pipeline of LLM-based agents. 
(d) shows the intermediate response returned to the user and the update of the experimental knowledge base. 
(e) shows the final results: the designed GNN, performance, and resource usage.}
    \vspace{-0.4cm}
    \label{fig1}
    \end{figure*}

\vspace{-0.2cm}
\subsection{Knowledge Bases Construction and Utilization}
\noindent
\textbf{Knowledge Bases Construction}
\label{sec-knowledge-base}
LLMs face challenges due to outdated knowledge and hallucinations. 
We address this by creating two knowledge bases, 
which is currently lacking for designing GNN model architectures.   
We collect resources and use the Knowledge Agent to manage them. 

The Knowledge Agent is tasked with acquiring and integrating 
specialized knowledge tailored to specific user requirements. 
This agent mainly manages two types of knowledge bases, as shown in Figure \ref{fig1}: 
the prior knowledge base and the experiment knowledge base. 
The prior knowledge base is enriched 
with task-specific information 
extracted from sources 
such as the Open Graph Benchmark (OGB) leaderboards, 
the PyTorch Geometric (PyG) documentation, 
and the top-tier conference proceedings that are accessible on Arxiv, 
ensuring the agent remains at the cutting edge of technology and methodology. 
The experiment knowledge base archives detailed experimental outcomes such as the benchmark evaluation results, 
including models setups and their performance on specific datasets, 
thereby providing insights into their effectiveness and application contexts.

The content of papers and reports often overlaps, 
with redundant background information and methods 
that can introduce noise and reduce the informativeness of retrieved knowledge. 
To address this, we employ a two-level knowledge extraction strategy, 
first, we start by summarizing inputs to obtain coarse-grained knowledge, 
then refine this into fine-grained details specific to graph learning tasks, 
such as architecture design and dataset usage. 
The code and the extended version with more details are available.
~\footnote{\url{https://github.com/lgssstsp/LLMNet}}.

\noindent
\textbf{Knowledge Utilization and Update}
To effectively utilize the constructed knowledge bases, 
we implement a goal-aware knowledge retrieval mechanism. 
Utilizing the RAG technique, 
we enhance the effectiveness of the designing GNN model architectures by retrieving relevant knowledge. 
The pre-trained model all-MiniLM-L6-v2 
encodes both the extracted knowledge and the queries from other agents. 
We calculate the cosine similarity in the embedding space to identify the most relevant knowledge. 
To accommodate the varying goals and resource types in graph learning, 
we apply a post-ranking strategy. 
The top-$k$ knowledge items from each resource type are initially retrieved 
and then re-ranked and selected by the knowledge agent based on the query's context. 
This refined knowledge is integrated into the graph learning agent's prompt, 
facilitating the design of GNN model.

LLMNet also incorporates a dynamically knowledge update mechanism. 
After the evaluation of a GNN model, 
the experimental summary, including the task plan, designed GNNs, and results, is stored in memory. 
The planning agent then compiles a report, 
which is added to the knowledge base, ensuring that the system's knowledge remains 
current and applicable for future pipeline runs. 
This continuous update process allows LLMNet to adapt and improve over time, 
enhancing its ability to design effective GNN models.


\vspace{-0.2cm}
\subsection{Knowledge-Guided GNNs Model Designing}

\begin{table*}[ht]
    \tiny
    \centering
    \setlength{\tabcolsep}{2pt}

    \begin{tabular}{c|ccccc|ccc|c|c}
        \toprule
        & Cora                            & Photo                           & Actor                           & Genius                          & obgn-arxiv                      & DD                              & Proteins                        & ogbg-molhiv                     & Amazon-Sports$(\downarrow)$        & Avg. Rank \\ \hline
        LLMNet      & \cellcolor{gray!30} 87.10(0.36) & \cellcolor{gray!30} 96.11(0.33) & \cellcolor{gray!30} 40.93(0.35) & \cellcolor{gray!30} 90.89(0.11) & \cellcolor{gray!30} 72.70(0.54) & \cellcolor{gray!30} 78.27(2.57) & \cellcolor{gray!30} 75.44(0.93) & \cellcolor{gray!30} 74.27(1.54) & \cellcolor{gray!30} 0.9298(0.0071) & 1      \\
        LLMNet (GL) & \underline{86.68(0.40)}                 & \underline{95.50(0.21)}                  & 39.59(0.39)                     & 90.33(0.15)                     & 72.30(0.54)                     & \underline{77.69(2.24)}                   & \underline{74.88(1.16)}                 & 73.37(1.23)                     & 0.9622(0.0103)                     & 2.5      \\ \midrule
        GCN           & 85.68(0.61)                     & 93.13(0.27)                     & 33.98(0.76)                     & 89.10(0.13)                     & 71.74(0.29)                     & 73.59(4.17)                     & 74.84(3.07)                     & \underline{73.89(1.46)}                 & 1.0832(0.0077)                     & 5.25      \\
        SAGE          & 86.18(0.35)                     & 94.60(0.25)                     & 39.28(0.18)                     & 89.71(0.09)                     & 71.49(0.27)                     & 76.99(2.74)                     & 73.87(2.42)                     & 73.46(1.69)                     & 0.9900(0.0125)                     & 4.5      \\ \midrule
        AutoML        & 86.57(0.32)                     & 95.38(0.30)                     & \underline{40.39(0.03)}                & \underline{90.81(0.04)}                 & \underline{72.42(0.37)}                 & 77.03(2.48)                     & 74.58(2.61)                     & 73.51(3.21)                     & \underline{0.9327(0.0006)}                 & 2.63      \\ \midrule
        LLM-GNN       & 84.64(1.04)                     & 93.73(0.38)                     & 38.92(0.07)                     & 89.31(0.17)                     & 70.83(0.93)                     & 75.12(3.44)                     & 74.47(3.65)                     & 72.93(0.90)                     & 0.9670(0.0079)                     & 5.13     \\  \bottomrule
    \end{tabular}
    \vspace{-0.2cm}
    \caption{Performance comparisons of the proposed LLMNet and baselines on three tasks. 
        We report the test accuracy and the standard deviation for node and graph classification tasks, and use the common Rooted Mean Square Error (RMSE) for the item ranking task.  
        The top-ranked performance in each dataset is highlighted in gray, and the second best one is underlined.
        The average rank on all datasets is provided in the last column.
    }
    \label{tb-comprehensive-perf}
    \vspace{-0.5cm}
\end{table*}

Figure \ref{fig1} illustrates how each agent engages 
with knowledge bases to streamline the entire process. 
The two knowledge bases bridge research and application, 
they empower agents to make informed decisions.

\noindent
\textbf{Planning Agent}
The Planning Agent generate a task plan based on user instructions,
to direct subsequent agent actions, 
which includes specifications for 
datasets, task types and evaluation metrics. 
After all agents completed their tasks, 
this agent evaluates the experimental results, 
utilizing insights from the experiment knowledge base to 
determine whether a revision loop is necessary.


\noindent
\textbf{Data Agent}
The Data Agent utilizes insights from the prior knowledge base 
to perform feature engineering tailored to specific graphs and tasks, 
ensuring alignment with expert practices in a knowledge-guided manner.

\noindent
\textbf{Configuration Agent}
The Configuration Agent is responsible for configuring 
the search space, 
which includes possible model architecture configurations 
such as layers and connections, 
and the search algorithm that explores this space.
It interacts with the prior knowledge base 
to gain insights on model design, 
enhancing the effectiveness of 
search space configuration and algorithm selection.

\noindent
\textbf{Evaluation Agent}
The Evaluation Agent is designed to finetune the designed GNN and conduct experiments 
to validate its performance. 
After completing the experiments, the Evaluation Agent 
transmits the results to the Knowledge Agent 
for integration into the experiment knowledge base.


\vspace{-0.2cm}

\section{Experiments}
We evaluate LLMNet's effectiveness on twelve datasets across three tasks as shown in Table~\ref{tb-comprehensive-perf}, 
the performance of another three datasets are shown in appendix of extended version.
Detailed resource costs and ablation studies are in the appendix of the extended version.

\vspace{-0.2cm}
\subsection{Experimental Settings}
\noindent\textbf{Datasets} We evaluate twelve widely used datasets 
across three tasks as shown in Table~\ref{tb-comprehensive-perf}. 
The detailed introduction of these datasets and 
the evaluation performance of another three datasets are shown 
in appendix of extended version.

\noindent\textbf{Baselines}
In this paper, we provide several kinds of baselines. 
(1) GNNs with task adaption, including GCN~\cite{kipf2016semi} and GraphSAGE~\cite{hamilton2017inductive} with task-specific adaptations.
(2) AutoML-based methods. We adopt F2GNN~\cite{wei2022designing} / LRGNN~\cite{wei2023search} / Prof-CF~\cite{wang2022automated} for three tasks.
(3) LLM-GNN. GNNs generated by LLMs.
(4) LLMNet (GL) operates without external knowledge.



\vspace{-0.2cm}
\subsection{Performance Comparisons}

Table~\ref{tb-comprehensive-perf} showcases the performance of LLMNet on twelve datasets across three tasks. 
LLMNet consistently outperforms all baselines, 
highlighting its ability to design effective GNNs for 
various graph learning tasks. 
The enhanced performance of LLMNet over LLMNet (GL) underscores 
the value of incorporating extracted knowledge into the GNN design process. 
Unlike AutoML methods that operate within a predefined design space, 
LLMNet (GL) leverages LLMs to expand this space, 
achieving comparable performance and validating the agents' problem-solving capabilities. 
The LLM-GNN baseline, which relies solely on LLM suggestions without knowledge integration, 
faces challenges in understanding tasks and graphs, 
resulting in less effective GNN designs. 
LLMNet's superior performance highlights the significance 
of knowledge in designing effective GNNs.

\vspace{-0.2cm}
\section{Demonstration}
\label{sec-demo}
In this section, 
we demonstrate the use case of LLMNet on a real-world problem. 
For example, users aim to 
predict the category of articles within a citation network.

\begin{figure}[t]
    \centering
    \includegraphics[width=1.00\columnwidth]{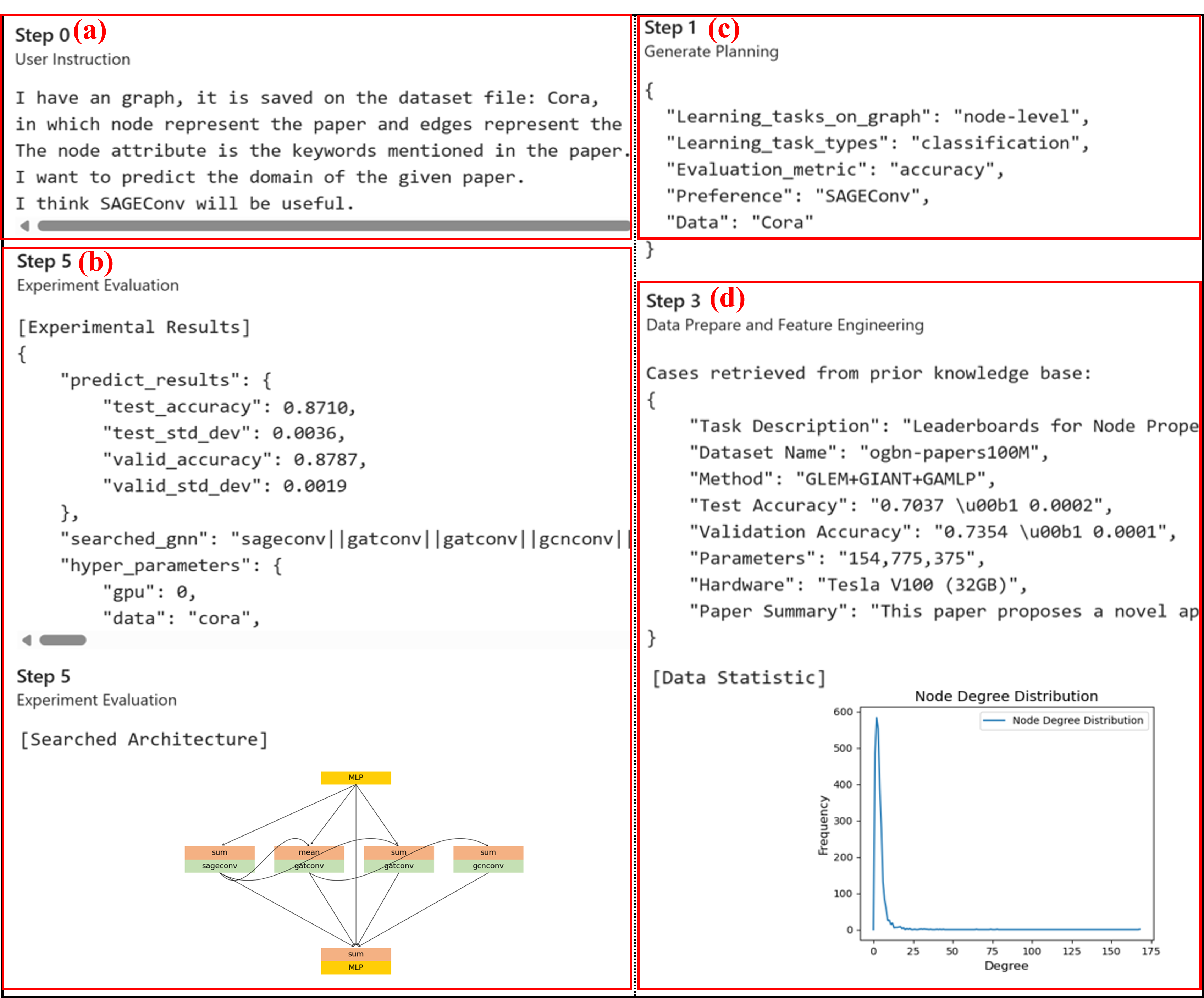}
    \caption{
		The detailed steps and output of LLMNet.
    }
    \vspace{-0.4cm}
    \label{fig2}
    \end{figure}

As shown in Figure~\ref{fig2}, 
\textbf{(a)} illustrates the user's input instructions, 
\textbf{(b)} displays the system's experimental results and its designed GNN model, 
LLMNet achieves an accuracy of 0.8710 on the Cora dataset, 
surpassing the GNN-based baselines 
GCN~\cite{kipf2016semi} at 0.8568, 
ACM-GCN~\cite{luan2022revisiting} at 0.8667 (Detailed experiments is in the extended version), 
and the AutoML-based baseline SANE~\cite{zhao2021search} at 0.8640. 
\textbf{(c)} displays the task plan generated by the Planning Agent, 
which interprets the user's intention to predict the category of articles within a citation network 
as a node classification task. 
\textbf{(d)} shows the Data Agent retrieving relevant knowledge 
from the prior knowledge base, including methods for node classification. 
It also visualizes graphs to better understand the data structure.


This demonstration showcases the effectiveness of LLMNet 
in automatically designing GNN model for real-world graph learning problems.

\section*{Acknowledgement}

This work is supported by 
National Key Research 
and Development Program of China (under Grant No.2023YFB2903904), 
the National Natural Science Foundation of China (under Grant No.~92270106), 
and the Beijing Natural Science Foundation (under Grant No.~4242039).

\bibliographystyle{named}
\bibliography{ref}

\clearpage

\end{document}